\documentclass[10pt,twocolumn,letterpaper]{article}

\usepackage{cvpr}              

\usepackage{graphicx}
\usepackage{amsmath}
\usepackage{amssymb}
\usepackage{booktabs}
\usepackage{amssymb}
\usepackage{pifont}
\newcommand{\cmark}{\ding{51}}%
\newcommand{\xmark}{\ding{55}}%
\usepackage{multirow}
\usepackage{makecell}
\usepackage{wrapfig}
\usepackage{footmisc}
\usepackage{cuted}
\usepackage[toc,page]{appendix}

\DeclareMathOperator*{\argmin}{arg\,min}
\usepackage[export]{adjustbox}[2011/08/13]
\usepackage[pagebackref,breaklinks,colorlinks]{hyperref}

\usepackage[capitalize]{cleveref}
\crefname{section}{Sec.}{Secs.}
\Crefname{section}{Section}{Sections}
\Crefname{table}{Table}{Tables}
\crefname{table}{Tab.}{Tabs.}

\def\cvprPaperID{3475} 
\def\confName{CVPR}
\def\confYear{2023}

\begin{document}

\title{Neural Vector Fields: Implicit Representation by Explicit Learning}

\author{Xianghui Yang\textsuperscript{1}, Guosheng Lin\textsuperscript{2}, Zhenghao Chen\textsuperscript{1}, Luping Zhou\textsuperscript{1}\\
\textsuperscript{1}The University of Sydney, \textsuperscript{2}Nanyang Technological University\\
{\tt\small \{xianghui.yang,zhenghao.chen,luping.zhou\}@sydney.edu.au, gslin@ntu.edu.sg}
}
\maketitle

\begin{abstract}
  Deep neural networks (DNNs) are widely applied for nowadays 3D surface reconstruction tasks and such methods can be further divided into two categories, which respectively warp templates explicitly by moving vertices or represent 3D surfaces implicitly as signed or unsigned distance functions. Taking advantage of both advanced explicit learning process and powerful representation ability of implicit functions, we propose a novel 3D representation method, Neural Vector Fields (NVF). It not only adopts the explicit learning process to manipulate meshes directly, but also leverages the implicit representation of unsigned distance functions (UDFs) to break the barriers in resolution and topology. Specifically, our method first predicts the displacements from queries towards the surface and models the shapes as \textit{Vector Fields}. Rather than relying on network differentiation to obtain direction fields as most existing UDF-based methods, the produced vector fields encode the distance and direction fields both and mitigate the ambiguity at ``ridge" points, such that the calculation of direction fields is straightforward and differentiation-free. The differentiation-free characteristic enables us to further learn a shape codebook via Vector Quantization, which encodes the cross-object priors, accelerates the training procedure, and boosts model generalization on cross-category reconstruction. The extensive experiments on surface reconstruction benchmarks indicate that our method outperforms those state-of-the-art methods in different evaluation scenarios including watertight vs non-watertight shapes, category-specific vs category-agnostic reconstruction, category-unseen reconstruction, and cross-domain reconstruction. Our code is released at~\url{https://github.com/Wi-sc/NVF}.
\end{abstract}


\begin{figure}[t]
  \centering
   \includegraphics[width=0.92\linewidth]{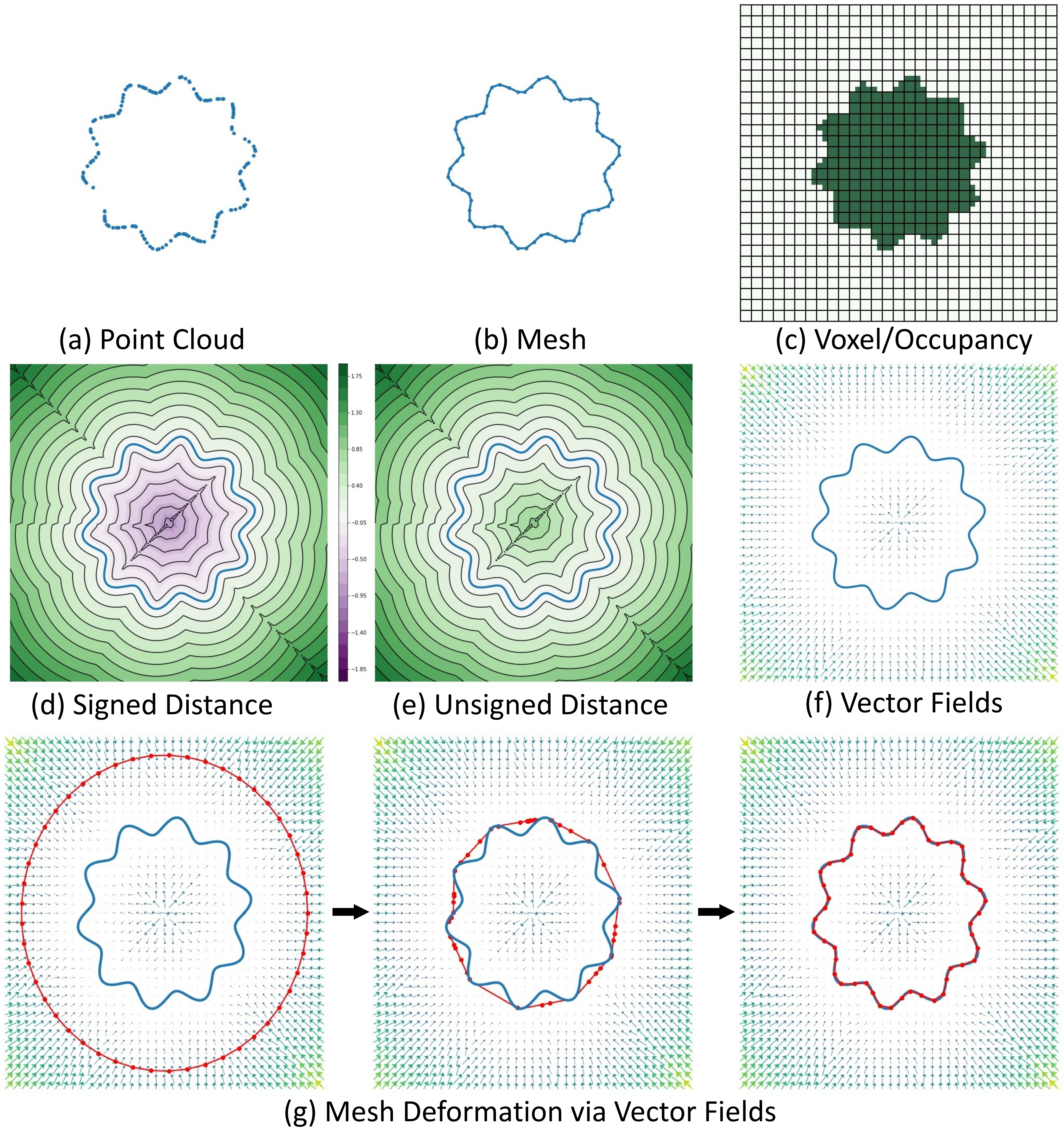}
   \vspace{-3mm}
   \caption{Common 3D representations. Explicit representations: (a) point clouds, (b) meshes, (c)  voxels. Implicit representations: (c)  occupancy, (d) reconstruction from the signed distance functions, and (e) reconstruction from unsigned distance functions. Our method represents continuous surfaces through (f) vector fields. (g) Vector fields can deform meshes (red) as explicit representation methods.}
   \label{fig:head}
\vspace{-3mm}
\end{figure}
\section{Introduction}
\label{sec:intro}

Reconstructing continuous surfaces from unstructured, discrete and sparse point clouds is an emergent but non-trivial task in nowadays robotics, vision and graphics applications, since the point clouds are hard to be deployed into the downstream applications without recovering to high-resolution surfaces~\cite{marching_cube, poisson, bspline,bpa}. 

\begin{figure}
  \centering
  \begin{subfigure}{0.4\linewidth}
    \includegraphics[width=1\linewidth]{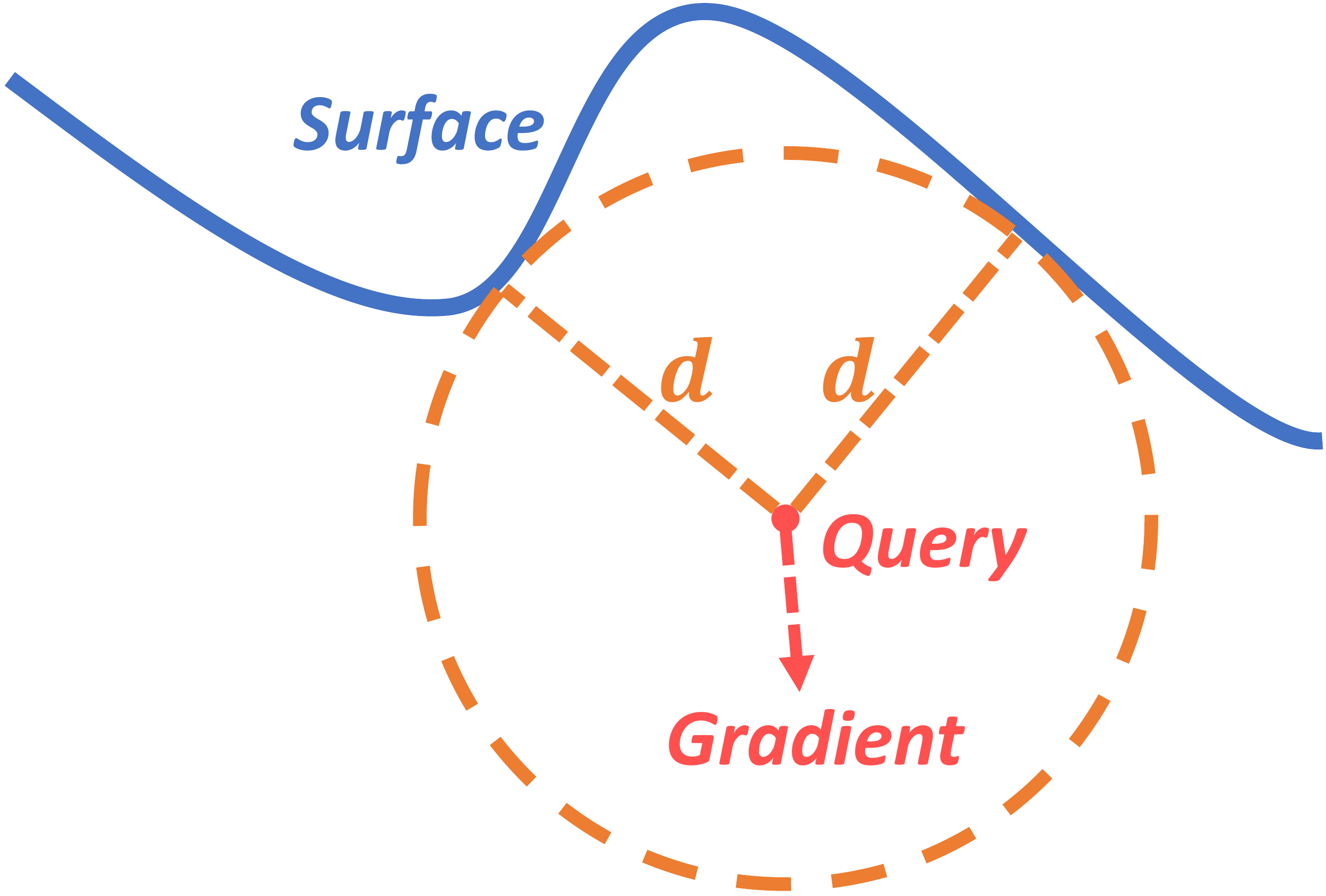}
    \caption{NDF~\cite{ndf}.}
    \label{fig:ambiguity-a}
  \end{subfigure}
  \begin{subfigure}{0.4\linewidth}
    \includegraphics[width=1\linewidth]{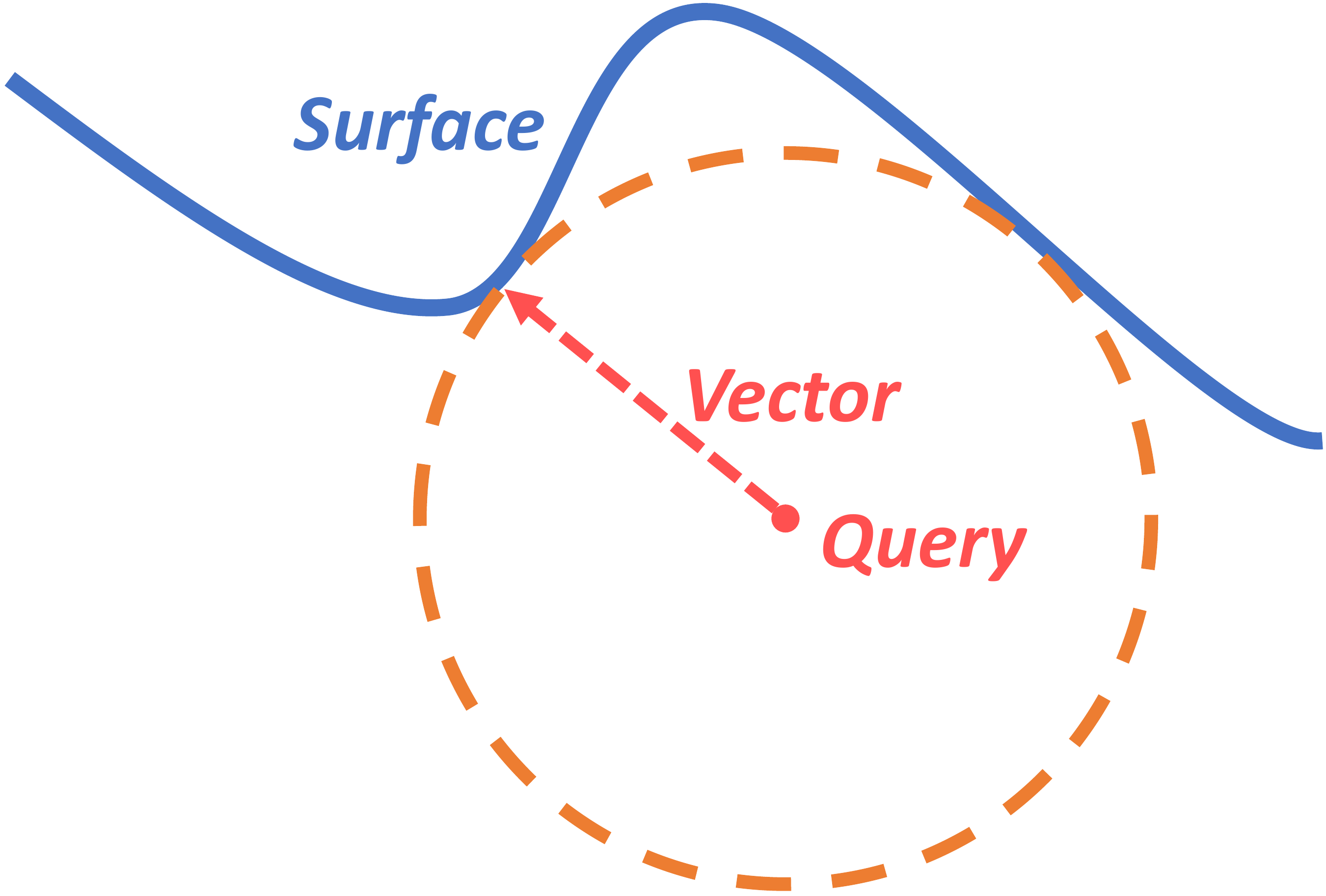}
    \caption{NVF.}
    \label{fig:ambiguity-b}
  \end{subfigure}
  \vspace{-3mm}
  \caption{Gradient ambiguities. (a) NDF~\cite{ndf} cannot guarantee to pull points onto surfaces (\ie, ambiguity of gradient), while (b) our NVF address this issue by direct displacement learning.}
  \label{fig:ambiguity}
  \vspace{-2mm}
\end{figure}

With the tremendous success of Deep Neural Networks (DNNs), a few DNN-based surface reconstruction methods have already achieved promising reconstruction performance. These methods can be roughly divided into two categories according to whether their output representations are explicit or implicit.
As shown in~\cref{fig:head}, explicit representation methods including \textit{mesh} and \textit{voxel} based ones denote the exact location of a surface, which learn to warp templates~\cite{atlasnet,atlasnet++,nmf,dsp,dgp,Meshlet} or predict voxel grids~\cite{octree,hierarchical,scalable}. Explicit representations are friendly to downstream applications, but they are usually limited by resolution and topology. 
On the other hand, implicit representations such as \textit{Occupancy} and \textit{Signed Distance Functions (SDFs)} represent the surface as an isocontour of a scalar function, which receives increasing attention due to their capacity to represent surfaces with more complicated topology and at arbitrary resolution~\cite{points2surf,occnet,ifnet,conv_occnet,local_implicit_grid,learning_implicit_fields}. However, most implicit representation methods usually require specific pre-processing to close non-watertight meshes and remove inner structures.
To free from the above pre-processing requirements for implicit representation, Chibane \etal~\cite{ndf} introduced~\textit{Neural Unsigned Distance Fields (NDF)},
which employs the~\textit{Unsigned Distance Functions (UDFs)} for neural implicit functions (NIFs) and models continuous surfaces by predicting positive scalar between query locations and the target surface. Despite certain advantages, UDFs require a more complicated surface extraction process than other implicit representation methods (\eg, SDFs). Such a process using Ball-Pivoting Algorithm~\cite{bpa} or gradient-based Marching Cube~\cite{meshudf,CAP_UDF} relies on model differentiation during inference (\ie, differentiation-dependent). Moreover, UDFs leave gradient ambiguities at “ridge” points, where the gradients\footnote{\label{gradient}Learning-based methods calculate the gradients of distance fields via model differentiation. The opposite direction of gradients should point to the nearest-neighbor point on the target surface.} used for surface extraction cannot accurately point at target points as illustrated by~\cref{fig:ambiguity-a}.



In this work, we propose a novel 3D representation method, \textit{Neural Vector Fields (NVF)}, which leverages the explicit learning process of direct manipulation on meshes and the implicit representation of UDFs to enjoy the advantages of both approaches. That is, NVF can directly manipulate meshes as those explicit representation methods as~\cref{fig:head}\textcolor{red}{g}, while representing the shapes with arbitrary resolution and topology as those implicit representation methods.
Specifically, NVF models the 3D shapes as vector fields and computes the displacement between a point $\mathbf{q}\in\mathbb{R}^3$ and its nearest-neighbor point on the surface $\mathbf{\hat{q}}\in\mathbb{R}^3$ by using a learned function $f(\mathbf{q})=\mathbf{\Delta q}=\mathbf{\hat{q}}-\mathbf{q}:\mathbb{R}^3\Rightarrow\mathbb{R}^3$. 
Therefore, NVF could serve both as an implicit function and an explicit deformation function, since the displacement output of the function could be directly used to deform source meshes (\ie,~\cref{fig:head}\textcolor{red}{g}). 
In general, it encodes both distance and direction fields within vector fields, which can be straightforwardly obtained from the vector fields. 

Different from existing UDF-based methods, our NVF representation avoids the comprehensive inference process by skipping the gradient calculation during the surface extraction procedure\footref{gradient}, and mitigates ambiguities by directly learning displacements as illustrated by~\cref{fig:ambiguity-b}. Such one-pass forward-propagation nature frees NVF from differentiation dependency, significantly reduces the inference time and memory cost, and allows our model to learn a shape codebook consisting of un-differentiable discrete shape codes in the embedded feature space. The learned shape codebook further provides cross-object priors to consequently improve the model generalization on cross-category reconstruction, and accelerates the training procedure as a regularization term during training. \textbf{We use VQ as an example to demonstrate that the differentiation-free property of NVF provides more flexibility in model design in this paper.} 


We conduct extensive experiments on two surface reconstruction benchmark datasets: a synthetic dataset ShapeNet~\cite{shapenet} and a real scanned dataset MGN~\cite{mgn}. Besides category-specific reconstruction~\cite{ndf,gifs} as demonstrated in most reconstruction methods, we also evaluate our framework by category-agnostic reconstruction, category-unseen reconstruction, and cross-domain reconstruction tasks to exploit the model generalization. Our experimental results indicate that our NVF can significantly reduce the inference time compared with other UDF-based methods as we avoid the comprehensive surface extraction step and circumvent the requirement of gradient calculation at query locations. Also, using the shape codebook, we observe a significant performance improvement and a better model generalization across categories.

Our contributions are summarized as follows.
\begin{itemize}
  \item {We propose a 3D representation \textit{NVF} for better 3D field representation, which bridges the explicit learning and implicit representations, and benefits from both of their advantages. Our method can obtain the displacement of a query location in a differentiation-free way, and thus it significantly reduces the inference complexity and provides more flexibility in designing network structures which may include non-differentiable components.}
    
  \item {
  Thanks to our differentiation-free design, we further propose a learned shape codebook in the feature space, which uses VQ strategy to provide cross-object priors. In this way, each query location is encoded as a composition of discrete codes in feature space and further used to learn the NVF.}

  \item {We conduct the extensive experiments to evaluate the effectiveness of our proposed method. It consistently shows promising performance on two benchmarks across different evaluation scenarios: water-tight vs non-water-tight shapes, category-specific vs category-agnostic reconstruction, category-unseen reconstruction, and cross-domain reconstruction.}

\end{itemize}
\section{Related Work}

\subsection{Explicit 3D Representations}
Early shape representation methods are built upon voxels~\cite{voxNet,pix3d,3dr2n2}. As the 3D spaces are discretized into girds, so they can be processed by adapting learning-based image processing techniques~\cite{surfacenet,stereo_machine,UnsupervisedLO,LearningAP,marrnet}. However, voxels are usually with the memory footprint scales cubically with the resolution. So recent voxel-based methods introduce adaptive discretization~\cite{octree,hierarchical,scalable,octnet} and voxel block hashing~\cite{volume_mapping} to alleviate higher computational requirements for higher resolution. 

Differently, point cloud based methods produce a compact and sparser encoding of surfaces, with lower computational cost and higher accessibility. Several learning-based point cloud processing methods ~\cite{pointnet,pointnet++,kpconv,point_transformer,edge_conv} are proposed in recent years and achieve tremendous success on 3D shape analysis ~\cite{deep_sets,pointwise,Relation-Shape,RandLA,Deep_Parametric,Minkowski,VoxelNet} and synthesis~\cite{pc_generation,FoldingNet,atlasnet,PCN}. %
However, point clouds usually lack rich geometry information (\eg, surfaces and topology), which results in limited applications in downstream tasks. 

To complete such geometry information, polygonal meshes~\cite{pixel2mesh,zeromesh,learning_category,photometric,Mesh_Autoencoders} define 3D shapes by graphs consisting of vertices and edges. However, they are usually suffered from fixed topology as most of them deform a template (\eg, ellipsoid~\cite{pixel2mesh}, sphere~\cite{tmn,learning_category,zeromesh}) into target shapes by moving vertices. Recent works addressed these issues by deforming from charts~\cite{atlasnet}, voxels~\cite{meshrcnn}, and pruning needless faces~\cite{tmn}. 

We observe that these methods encourage the networks to simulate vector fields within 3D space. Inspired by them, we learn the model using explicit learning with the vector fields instead of mesh-to-mesh distances to learn vector fields. In this way, we can break the boundaries of fixed topology and resolution of templates.

\subsection{Implicit 3D Representations}
To overcome the limited resolution and fixed mesh topology barriers from most explicit representation, implicit representation methods represent continuous surfaces through implicit functions. Such representations are usually implemented by multi-layer perceptions (MLP), which generate binary occupancy~\cite{occnet, ldif, ifnet, NASA, PIFu, PIFuHD} and SDFs~\cite{deepsdf,local_implicit_grid,learning_implicit_fields} given locations as queries. 
On the other hand, those methods require a heavy pre-processing stage to close the shapes artificially to obtain watertight meshes due to the characteristics of occupancy and SDFs. For one thing, It is also non-trivial to define the inside and outside of open surfaces. For another, the pre-processing results in substantive geometry information loss and lack of generalization on non-watertight meshes.

To model general non-watertight shapes, Chibane \etal~\cite{ndf} introduced UDFs to learn the unsigned distance.
Recently, other methods improve the performance and generalization of UDFs~\cite{anchor,csp,rangeudf,gifs,CAP_UDF}. For example, GIFS~\cite{gifs} adopts an intersection classification branch and CAP-UDF~\cite{CAP_UDF} directly optimizes models on raw point clouds. Although such UDF-based methods can intuitively improve the generalization of implicit representation, they require non-trivial surface extraction processing such as Ball-Pivoting algorithm~\cite{ndf} and gradient-based Marching Cubes~\cite{meshudf,CAP_UDF}.

Different than aforementioned UDF-based methods, our NVF does not rely on differentiation to obtain gradients as direction fields and can simply obtain distance and direction fields by one-time forward propagation. Such effective differentiation-free strategy can significantly reduce the inference burden.

\subsection{Vector Quantization}
The Vector Quantization (VQ) strategy converts the extracted features into quantized and compact latent representations. Oord \etal first introduced VQ-VAE~\cite{vq-vae} to combine VQ strategy with Variational AutoEncoder for images and speech generation. 
Then, Yu \etal introduce VQ-GAN~\cite{vqgan} to cooperate codebooks with adversarial learning for HQ image synthesis.
Overall, several methods employ this strategy for image generation~\cite{Autoregressive_image,vq-vqe2,CogViewMT,VectorQD}, speech synthesis~\cite{Audio-Visual,jukebox}, video~\cite{CogVideo,PredictingVW,LongVG,DLFormerDL}, shape generation~\cite{AutoSDF,autoregressive,3DILG,shapeFormer} and compression~\cite{Soft-to-Hard,Chen2020DifferentiablePQ}. In this work, we also employ VQ and adopt a multi-head codebook in feature space to denoise, accelerate training and improve generalization on shape reconstruction.

\begin{figure*}[h]
  \centering
   \includegraphics[width=0.99\linewidth]{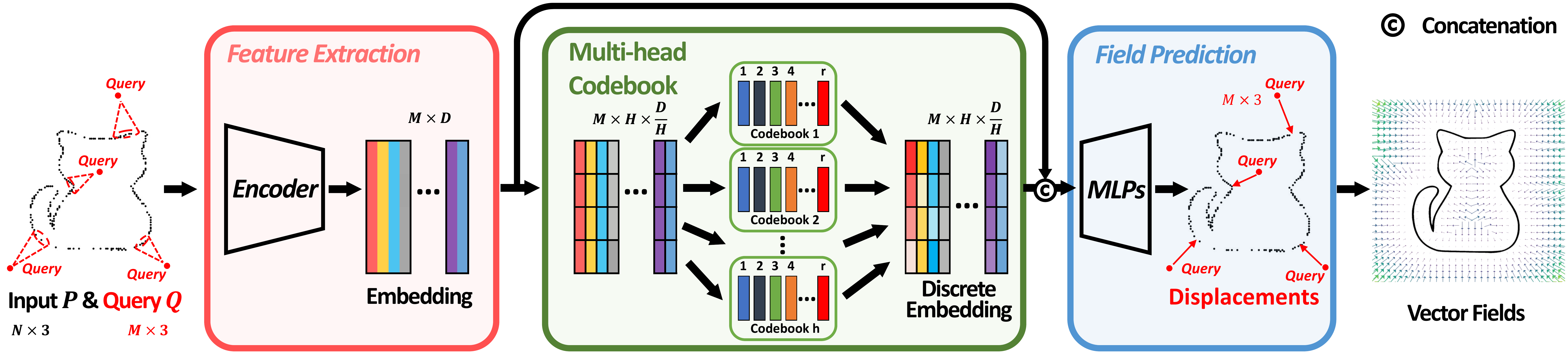}
   \vspace{-3mm}
   \caption{Overview of our NVF framework. It encodes the input point cloud (black dots), samples features for each query point (red dots) and discretizes query embeddings via Vector Quantization to predict the displacements of query location to the target surface (red arrow).}
   \label{fig:main}
   \vspace{-2mm}
\end{figure*}

\section{Methodology}

%
Given a 3D query location $\mathbf{q}$ from a query set $Q\in\mathbb{R}^{M\times 3}$ and a sparse point cloud $P\in\mathbb{R}^{N\times3}$, our NVF framework predicts the displacement vector $\Delta\mathbf{q}\in\mathbb{R}^3$ that moves $\mathbf{q}$ to the underlying surface of $P$. This is achieved through three main modules, \ie, Feature Extraction, Multi-head Codebook, and Field Prediction, as illustrated in~\cref{fig:main}. Once the predicted vector field is obtained, we adopt Marching Cubes to generate the surface mesh.
The details of each module are elaborated as follows.

\subsection{Neural Vector Fields} 
As mentioned above, our NVF is a neural network function, which models shapes $\mathcal{S}$ by predicting the field of displacements $\Delta\mathbf{q}$ for a given query point $\mathbf{q} \in Q$ to its nearest point $\mathbf{\hat{q}}\in\mathcal{S}$ on the surface.  
It is formulated by
\begin{equation}
    NVF(\mathbf{q})=\Delta\mathbf{q}={\min_{\hat{q}\in\mathcal{S}}}\ \mathbf{\hat{q}}-\mathbf{q}.
\end{equation}
Please note that NVFs are directional distance fields, which encode both distance fields $d=||\mathbf{\hat{q}}-\mathbf{q}||_2$ and direction fields $\mathbf{g}=(\mathbf{\hat{q}}-\mathbf{q})/||\mathbf{\hat{q}}-\mathbf{q}||_2$ in a straightforward way. Therefore, unlike distance fields, our NVF does not require the differentiation of distance fields for direction information. This property is referred to as ``differentiation-free" in this paper, which significantly reduces the inference time. 
On the other hand, our NVF enjoys the merits of UDFs and can represent much wider classes of surfaces and manifolds even not necessarily closed. 
%

\subsection{Feature Extraction}\label{sec:FeatureExtraction}
\label{shape_encoding}
The input point cloud $P$ is first sent to a feature encoder to extract 3D shape features for each point, which leads to the output feature maps $F\in \mathbb{R}^{N\times C}$. Multiple point cloud networks ~\cite{point_transformer,pointnet++,pointnet} could be used as the feature encoder, and an ablation study about the backbones is given in our experiment. For the query point $\mathbf{q}$, we find its $K$ nearest points ${\mathbf p}_i (i=1, \cdots, k)$ in $P\in\mathbb{R}^{N\times 3}$ based on a search using Euclidean distance, and form the feature embedding $z_q$ of $\mathbf{q}$ by concatenating the signatures $z_{qi}$ of its nearest points ${\mathbf{p}}_i$.  Each signature $z_{qi}$ is a non-linear mapping of the position features (\ie, the positions of the query point $\mathbf{q}$, the nearest point $\mathbf{p}_i$, and the relative position $\mathbf{q}-\mathbf{p_i}$) and the shape features (i.e., $\mathbf{p_i}$'s corresponding feature $f_i \in F$), which is realized by MLP. Mathematically, the query embedding $z_q\in\mathbb{R}^{D}$ is obtained by
\begin{equation}
\begin{aligned}
    &z_q=z_{q1}\oplus z_{q2}\oplus...z_{qi}...\oplus z_{qk}, \\
    &z_{qi}=MLP(\mathbf{q},\mathbf{p_i},\mathbf{p_i}-\mathbf{q},f_i), ~~~~ i=1,2,...,k.
\end{aligned}
\end{equation}
The symbol $\oplus$ indicates the concatenation operation. As can be seen, the query embedding $z_q$ carries the information of both the query point $\mathbf{q}$ and its $K$ nearest points on $P$.

\subsection{Multi-head Codebook} 
  
  After obtaining the embedding $z_q$ whose dimension is $D$, we learn a multi-head codebook $\mathcal{C}\in\mathbb{R}^{H\times R\times \frac{D}{H}}$ to produce cross-object priors. Our codebook contains $H$ sub-codebooks (i.e., heads). Each sub-codebook ${\mathcal C}^h$ ($h=1, \cdots, H$) contains $R$ discrete codes $c^h_r$ ($r=1, \cdots, R$) and each code has a dimension of $\frac{D}{H}$. The codebook is randomly initialized.

Accordingly, we split the continuous embedding $z_q$ into $H$ segments along the channel dimension. For each embedding segment $z_q^h$, we search its closest code from the corresponding sub-codebook ${\mathcal C}^h$ according to their Euclidean distance. In this way, we could discretize the continuous embedding $z_q$ to $\hat{z}_q$ that is composed of discrete segment codes. 
This process can be formulated as follows,
\begin{equation}
\begin{aligned}
 &z_q=z_q^{1}\oplus z_q^{2}\oplus...\oplus z_q^h...\oplus z_q^H,\\
    &\hat{z}_q=c^{1}_{r_1^*}\oplus c^{2}_{r_2^*}\oplus...\oplus c^{h}_{r_h^*}...\oplus c^{H}_{r_H^*}, \\
 & r_h^*=\argmin_{r\in\{1,...,R\}}||z_q^{h}-c^h_r||_2.
\end{aligned}
\end{equation}
Compared with using only one codebook $\mathcal{C}\in\mathbb{R}^{(H*R)\times D}$, the multi-head codebook enables $R^H$ permutations, which extends the feature space from $\mathbb{R}^{H*R}$ to $\mathbb{R}^{R^H}$, and enhances the codebook representation capacity. Note that the nearest search stops the gradient here. 

The multi-head codebook can be jointly learned during training by minimizing the distance between the continuous embedding $z$ and its discretization $\hat{z}_q$ as follows,
\begin{equation}
\label{commit}
    \mathcal{L}_{code}=\sum_{h\in\{1,...,H\}} ||sg(c^h_{r^*_h})-z_q^h||_2^2+\beta ||sg(z_q^h)-c^h_{r^*_h}||_2^2,
\end{equation}
where $sg$ stands for \textit{stop gradient} operation and $\beta$ is the weight to the commitment loss in the second term. Alternatively, we can update the discretized embedding $\hat{z}_q$ and therefore its corresponding codes ($\hat{z}_q^h \equiv c^{h}_{r_h^*}$) in the codebook via \textit{Exponential Moving Average}~\cite{vq-vae},
\begin{equation}
    c^h_{r^*_h} := \gamma c^h_{r^*_h} + (1-\gamma) z_q^h,
\end{equation}
with $\gamma$ is a value between 0 and 1. 

\subsection{Field Prediction}
Last, we take the continuous embedding $z_q$ and its discretization $\hat{z}_q$ to predict a vector as the displacement for the query point $\mathbf{q}$ to move to the ground truth surface $\mathcal{S}$. By combining the embeddings $z_q$ and $\hat{z}_q$, the final vector field is predicted using MLPs, i.e., $NVF(\mathbf{q})\approx \Delta \mathbf{q}=MLP(z_q,\hat{z}_q)$.

\subsection{Optimization}
\label{sec:optimization}




We optimize NVF by minimizing the difference between the predicted displacement $\mathbf{\Delta q}$ and the ground-truth displacement $\mathbf{\hat{q}-q}$, as well as the error of discretizing the continuous embedding $z_q$ into $\hat{z}_q$. We use stop-gradient operation (\ie, \textit{sg}) to cut the gradient after vector quantization. In sum, our overall objective function is,

\begin{equation}
    \mathcal{L}=\sum_{\mathbf{q}\in Q} ||\mathbf{\hat{q}}-\mathbf{q}-\mathbf{\Delta q}||_1 + \lambda \sum ||z_q-sg(\hat{z}_q)||_2^2.
\end{equation}

The hyperparameter $\lambda$ balances the two loss terms. The commitment loss term $||sg(z)-z^q||_2^2$ used in Eq.~\ref{commit} is replaced by the \textit{Exponential Moving Average} algorithm to update corresponding codes in the multi-head codebook.

\subsection{Surface Extraction}



Both signed and unsigned distance functions define the surfaces as a 0-level set. SDFs extract surfaces through Marching Cubes~\cite{marching_cube} faithfully due to the intersection being easily captured via sign flipping. Recently, MeshUDF~\cite{meshudf} and CAP-UDF~\cite{CAP_UDF} propose to detect the opposite gradient directions to replace the sign in SDFs and successfully apply the Marching Cubes~\cite{marching_cube} on UDFs. Note that these two methods learn distance fields and they need to differentiate the distance field to obtain the gradient direction. In contrast, our NVF can similarly extract surfaces using Marching Cubes while avoiding the differentiation of the distance field. This is because our NVF directly encodes both distance and direction fields. The surface extraction algorithm divides the space into grids and decides whether the adjacent corners locate on the same side or two sides. NVF predicts the vectors $\mathbf{\Delta q}$ of all lattices and normalizes them into normal vectors $\mathbf{g_i}=\Delta \mathbf{q_i}/||\Delta \mathbf{q_i}||_2$. Given a lattice gradient $\mathbf{g_i}$ and its adjacent lattices' gradients $\mathbf{g_j}$, the algorithm checks if their gradients have opposite orientations, and assigns the pseudo sign $s_i$ to the lattice. For more details, please refer to MeshUDF~\cite{meshudf}.


\begin{table}[t]
\centering
\resizebox{0.4\textwidth}{!}{
    \begin{tabular}{@{}c|c|c|c|c@{}}
    \hline
    Methods & CD$\downarrow$ & EMD$\downarrow$ & F1$_{1\times10^{-5}}$ & F1$_{2\times10^{-5}}$ \\
    \hline
    Input & 0.363 & 0.707 & 23.735 & 41.588 \\
    NDF~\cite{ndf} & 0.197 & 1.248 & 64.116 & 84.902 \\
    GIFS~\cite{gifs} & 0.146 & 0.970 & 54.867 & 79.722 \\
    \hline
     Ours & \textbf{0.114}& \textbf{0.945} & \textbf{64.261} & \textbf{85.290} \\
     \hline
    \end{tabular}
    }
    \vspace{-3mm}
    \caption{Quantitative evaluation on ShapeNet Cars. We train and evaluate our method on the raw data of the ShapeNet “Car” category. Our method achieves better performance than the state-of-the-art UDF-based methods.}
    \label{tab:car}
\end{table}

\begin{table*}[t]
\centering
\resizebox{0.85\textwidth}{!}{
    \begin{tabular}{@{}c|c|c|c|c|c|c|c|c@{}}
    \hline
    \multirow{2}{*}{Methods} & \multicolumn{4}{c|}{Base} & \multicolumn{4}{c}{Novel}\\
    \cline{2-9}
    & \multicolumn{1}{c|}{CD$\downarrow$} & \multicolumn{1}{c|}{EMD$\downarrow$} & \multicolumn{1}{c|}{F1$_{2.5\times10^{-5}}\uparrow$} & \multicolumn{1}{c|}{F1$_{1\times10^{-4}}\uparrow$}& \multicolumn{1}{c|}{CD$\downarrow$} & \multicolumn{1}{c|}{EMD$\downarrow$} & \multicolumn{1}{c|}{F1$_{1\times10^{-5}}\uparrow$} & \multicolumn{1}{c}{F1$_{2\times10^{-5}}\uparrow$} \\
    \hline
    Input & 0.840 & 1.045 & 14.148 & 25.111 & 0.800 & 1.024 & 17.576 & 29.815 \\
    OccNet~\cite{occnet} & 2.766 & 1.694 & 30.877 & 46.644 & 44.762 & 4.013 & 15.943 & 24.433 \\
    IF-Net~\cite{ifnet} & 0.190 & 1.120 & 65.975 & 85.421 & 0.596 & 1.608 & 61.670 & 81.106 \\
    NDF~\cite{ndf} & 0.169 & 1.538 & 66.802 & 84.809 & 0.169 & 1.741 & 65.622 & 84.069 \\
    GIFS~\cite{gifs} & 0.179 & 1.280 & 56.188 & 78.458 & 0.194 & 1.534 & 56.644 & 78.016 \\
    \hline
     Ours & ~\textbf{0.091} & ~\textbf{1.079} & ~\textbf{78.503} & ~\textbf{91.408} & ~\textbf{0.144} & ~\textbf{1.145} & ~\textbf{80.883} & ~\textbf{91.836} \\
     \hline
    \end{tabular}
    }
    \vspace{-3mm}
    \caption{Quantitative results of category-agnostic and category-unseen reconstructions on watertight shapes of ShapeNet. We train all models on the base classes, and evaluate them on the base and the novel classes, respectively.}
    \label{tab:watertight}
\end{table*}

\begin{table*}[t]
\centering
\resizebox{0.85\textwidth}{!}{
    \begin{tabular}{@{}c|c|c|c|c|c|c|c|c@{}}
    \hline
    \multirow{2}{*}{Methods} & \multicolumn{4}{c|}{Base} & \multicolumn{4}{c}{Novel}\\
    \cline{2-9}
    & \multicolumn{1}{c|}{CD$\downarrow$} & \multicolumn{1}{c|}{EMD$\downarrow$} & \multicolumn{1}{c|}{F1$_{2.5\times10^{-5}}\uparrow$} & \multicolumn{1}{c|}{F1$_{1\times10^{-4}}\uparrow$}& \multicolumn{1}{c|}{CD$\downarrow$} & \multicolumn{1}{c|}{EMD$\downarrow$} & \multicolumn{1}{c|}{F1$_{1\times10^{-5}}\uparrow$} & \multicolumn{1}{c}{F1$_{2\times10^{-5}}\uparrow$} \\
    \hline
    Input & 0.317 & 0.867 & 32.875 & 51.105 & 0.289 & 0.843 & 39.902 & 58.092 \\
    NDF~\cite{ndf} & 0.099 & 1.372 & 72.425 & 88.754 & 0.093 & 1.532 & 76.162 & 89.977 \\
    GIFS~\cite{gifs} & 0.118 & 1.260 & 64.915 & 85.115 & 0.296 & 1.499 & 69.252 & 86.518 \\
    \hline
    Ours & \textbf{0.085} & \textbf{1.197} & \textbf{75.372} & \textbf{90.266} & \textbf{0.078} & \textbf{1.340} & \textbf{79.723} & \textbf{91.576}\\
     \hline
    \end{tabular}
    }
    \vspace{-3mm}
    \caption{Quantitative results of category-agnostic and category-unseen reconstructions on non-watertight shapes of ShapeNet. We train all models on the base classes and evaluate them on the base and the novel classes, respectively.}
    \label{tab:non_watertight}
    \vspace{-2mm}
\end{table*}

\begin{table}[!h]
\centering
\resizebox{0.4\textwidth}{!}{
    \begin{tabular}{@{}c|c|c|c|c@{}}
    \hline
    Methods & CD$\downarrow$ & EMD$\downarrow$ & F1$_{1\times10^{-5}}$ & F1$_{2\times10^{-5}}$ \\
    \hline
    Input & 0.124 & 0.157 & 52.189 & 72.969 \\
    NDF~\cite{ndf} & 0.025 & 0.216 & 96.338 & 98.687 \\
    GIFS~\cite{gifs} & 0.039 & 0.192 & 93.330 & 97.295 \\
    \hline
     Ours & \textbf{0.014} & \textbf{0.184} & \textbf{98.499} & \textbf{99.498}\\
     \hline
    \end{tabular}
    }
    \vspace{-3mm}
    \caption{Quantitative results of cross-domain reconstruction on MGN~\cite{mgn}. We train our models based on ShapeNet with the base classes and evaluate them on the raw data from MGN.}
    \label{tab:garment}
    \vspace{-2mm}
\end{table}

\begin{figure}[h]
  \centering
  \includegraphics[width=0.95\linewidth]{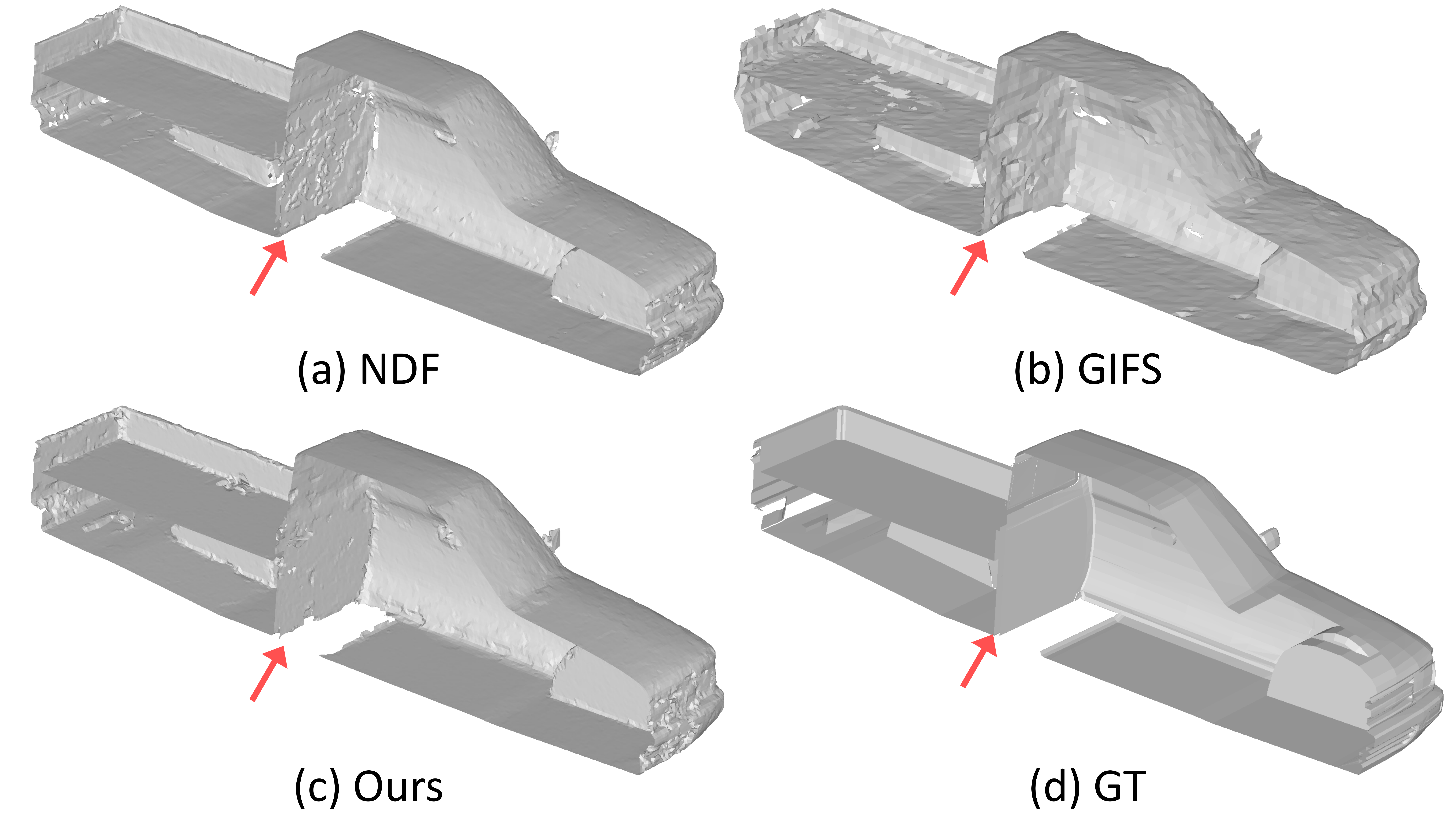}
  \vspace{-3mm}
   \caption{Qualitative visualization of Category-specific reconstruction on ShapeNet Cars. We cut parts of the shapes to visualize inner structures better.}
   \label{fig:car}
\end{figure}

\begin{figure*}[h]
  \centering
  \includegraphics[width=0.95\linewidth]{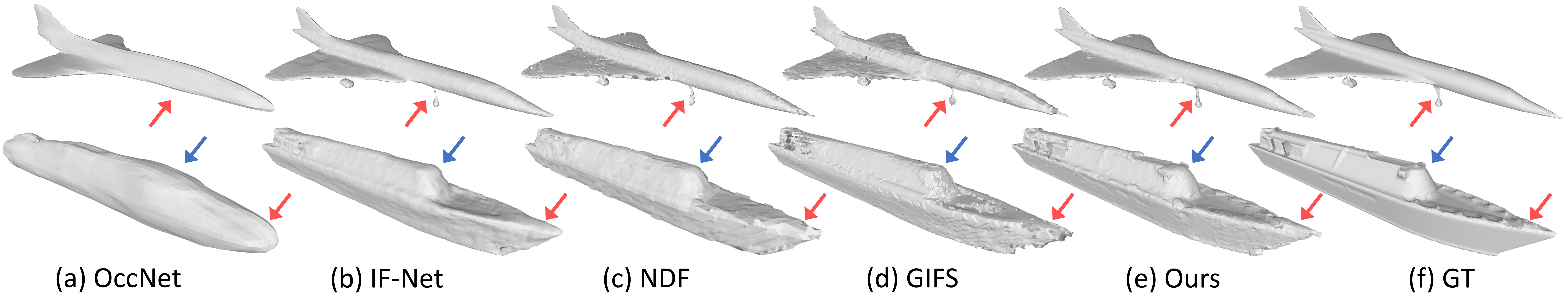}
  \vspace{-3mm}
   \caption{Visualization of category-agnostic and category-unseen reconstructions on watertight shapes from the ShapeNet dataset. The $1^{st}$ row, planes, is from the base classes. The $2^{rd}$ row, watercraft, is from the novel classes.}
   \label{fig:watertight}
\end{figure*}

\begin{figure}[h]
  \centering
  \includegraphics[width=0.95\linewidth]{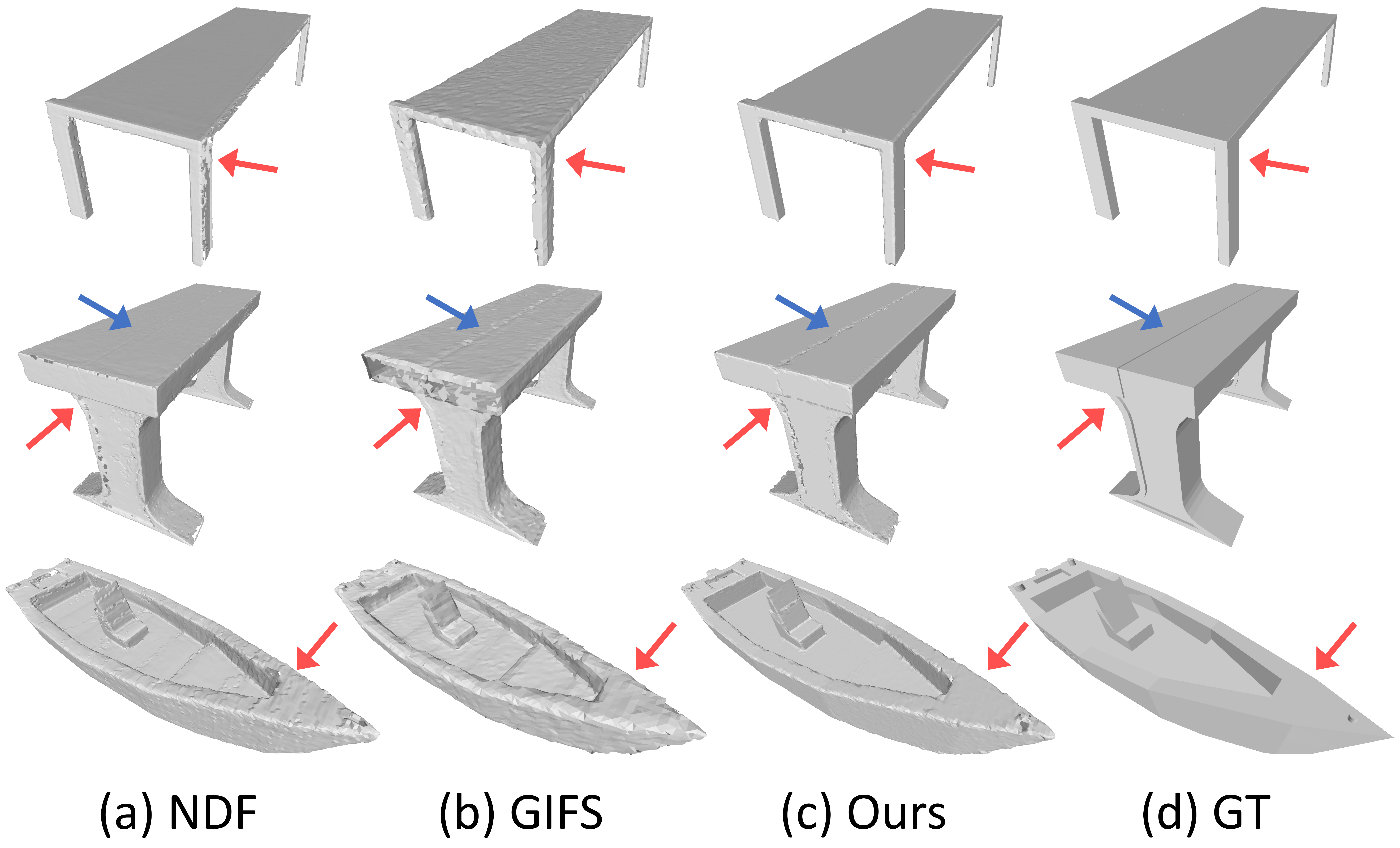}
  \vspace{-3mm}
   \caption{Visualization of category-agnostic and category-unseen reconstructions on non-watertight shapes from the ShapeNet dataset. The $1^{st}$ row, tables, is from the base classes. The $2^{nd}$ and $3^{rd}$ rows, benches and watercraft, are from the novel classes.}
   \label{fig:non_watertight}
\end{figure}

\begin{figure}[h]
  \centering
  \includegraphics[width=0.95\linewidth]{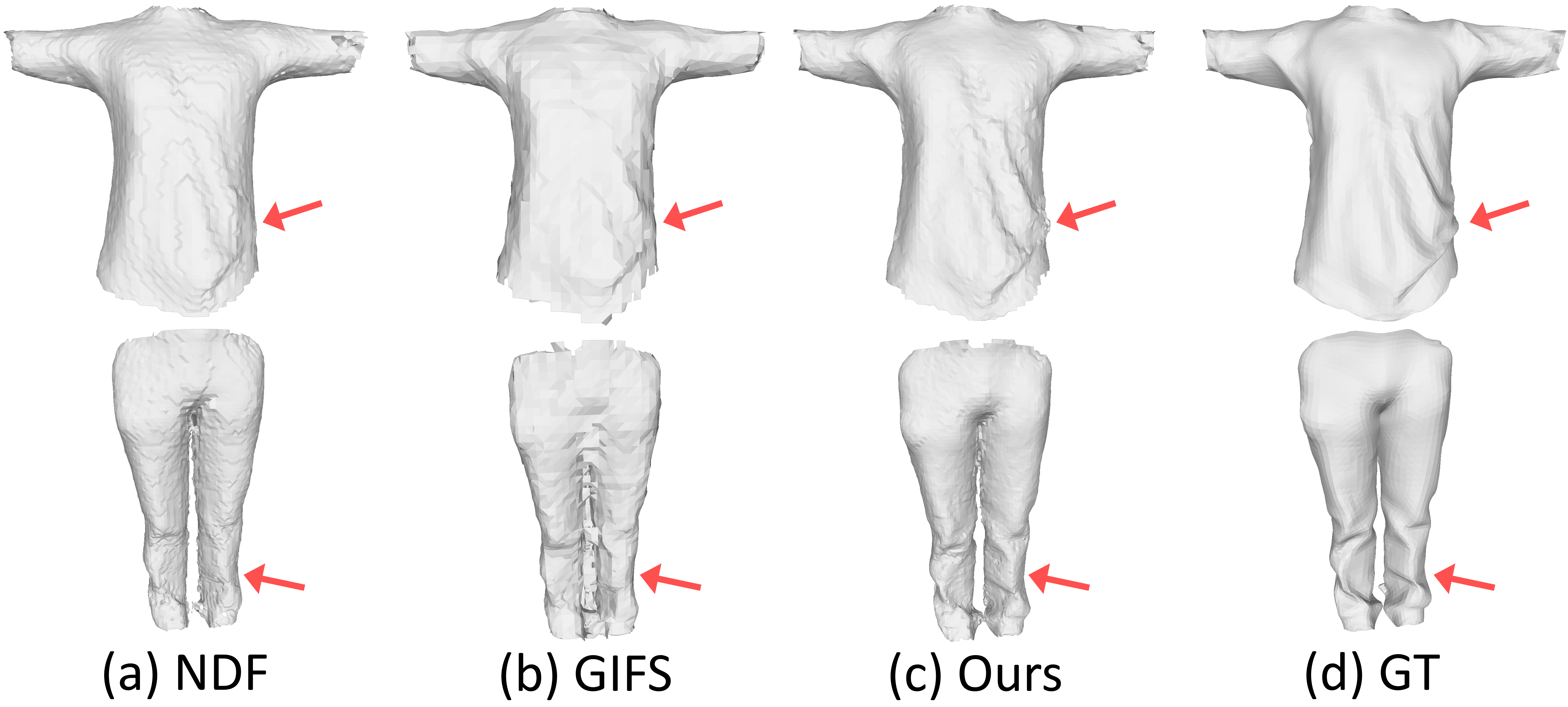}
  \vspace{-3mm}
  \caption{Visualization of cross-domain reconstruction on the MGN dataset. All models are trained based on ShapeNet with the base classes and evaluated directly on MGN.}
   \label{fig:garment}
\end{figure}


\section{Experiments}

\subsection{Experimental Protocol}
\noindent\textbf{Tasks.} We evaluate the effectiveness of our framework on four tasks: 1) category-specific, 2) category-agnostic, 3) category-unseen and 4) cross-domain reconstruction. We first demonstrate the ability of our NVF to reconstruct non-watertight meshes by category-specific reconstruction in~\cref{sec:specific}. Next, to evaluate the generalization ability, we compare our NVF with existing methods on category-agnostic and category-unseen reconstruction in ~\cref{sec:agnostic}. We also test cross-domain reconstruction by reconstructing real scanned data without training or fine-tuning in~\cref{sec:cross}.



\noindent\textbf{Implementations.} We employ PointTransformer~\cite{point_transformer} as our feature encoder and set $k=16$ for the nearest points. The multi-head codebook consists of 4 sub-codebooks, each containing 128 codes and 64 dimension channels. We set the grid resolution for surface extraction as $256$. To train our network, the learning rate is set as $l_r=0.001$ and decays with the factor $0.3$ for the subsequent 30, 70, and 120 steps. The loss weight $\lambda$ is set to be $0.001$. 

\noindent\textbf{Datasets.} Most experiments are conducted on a large-scale synthetic dataset ShapeNet~\cite{shapenet} and a real scanned dataset MGN~\cite{mgn}. We conduct our category-specific, category-agnostic, and category-unseen reconstruction experiments on ShapeNet~\cite{shapenet} while conducting cross-domain reconstruction experiments on MGN~\cite{mgn} to demonstrate our model generalization in the wild. All meshes are normalized to a unit cube with a range $[-0.5, 0.5]$.

ShapeNet is a synthetic dataset with 55 object categories. We choose \textit{cars} for category-specific reconstruction following UDF literature, and \textit{cars, chairs, planes, and tables} for category-agnostic reconstruction. For category-unseen reconstruction, we take \textit{speakers, bench, lamps, and watercraft} as the testing set. The above 4 categories for category-agnostic reconstruction are referred to as \textit{base classes} and the ones for category-unseen reconstruction are referred to as \textit{novel classes} in~\cref{tab:watertight} and~\cref{tab:non_watertight}.

MGN is a real scanned dataset containing 5 garment categories, (\ie, long coat, pants, shirt, short pants, and T-shirt), which are all open surfaces. We evaluate our model trained by ShapeNet on MGN to demonstrate the reconstruction ability in the wild. The results are in ~\cref{sec:cross}.

\noindent\textbf{Metrics.} 
We use Chamfer Distance (CD), Earth Mover Distance (EMD) and F-score as our evaluation metrics. Specifically, we sample 100k points from the reconstructed surfaces for the Chamfer Distance (CD $\times 10^{-4}$), 2048 points for the Earth Mover Distance (EMD $\times 10^{-2}$), and 100k points for F-score ($\times 10^{-2}$) with thresholds $1\times10^{-5}$ and $2\times10^{-5}$. 


\subsection{Category-specific Cases}
\label{sec:specific}
We first report the results of \textit{category-specific reconstruction} on ShapeNet cars to demonstrate the representation ability to one category of our NVF. We perform the same training/testing split and sample 10k points as the input as in NDF~\cite{ndf} and GIFS~\cite{gifs}. The quantitative comparison in~\cref{tab:car} shows that our method achieves better results with previous state-of-the-art methods including NDF and GIFS. Specifically, on CD and EMD, our NVF outperforms the second-best GIFS; and on F-scores, ours outperforms the second-best NDF. Moreover, we provide the qualitative comparison in~\cref{fig:car}. The visualization indicates that although NDF, GIFS, and our NVF are able to recover inner structures, our NVF could yield smoother surfaces and fewer holes in the finer parts (\eg, the back of driving seats).

\subsection{Category-agnostic and Category-unseen Cases}
\label{sec:agnostic}
\noindent\textbf{Base vs Novel categories on watertight shapes.}
We conduct category-agnostic and category-unseen reconstruction experiments to explore the generalization ability of our NVF. For a fair comparison with the previous methods (\eg, OccNet~\cite{occnet} and IF-Net~\cite{ifnet}) which cannot handle non-watertight shapes as the input, we report the quantitative experimental results on watertight shapes pre-processed by DISN~\cite{DISN} in~\cref{tab:watertight} and give qualitative visualization in~\cref{fig:watertight}. We use the same training/testing split as in IF-Net~\cite{ifnet}. The input of all methods is 3k points in this experiment.

~\Cref{tab:watertight} indicates that the surface reconstruction results produced by UDF-based methods are much better than occupancy representation, especially for novel classes. Among them, our NVF outperforms others by a large margin, no matter in base classes or novel classes.
The qualitative comparison in ~\cref{fig:watertight} shows that OccNet and IF-Net fail to construct a lot of details. NDF and GIFS are better than them, but they still fail to capture some details, \eg, undercarriages of planes, or little bumps from watercraft. In contrast, our method can reconstruct flat and faithful surfaces, and thereby achieves a better visual effect. 

\noindent\textbf{Base vs Novel categories on non-watertight shapes.} 
We also test on non-watertight shape reconstruction with the same setting above and the results are in~\cref{tab:non_watertight} and ~\cref{fig:non_watertight}. For the training/testing split, we randomly sample 3000 shapes from each bases class as the training set and 200 shapes from each base and novel class as the validation and testing set. The input is 10k points sampled from the raw data of these shapes.

The quantitative comparison in~\cref{tab:non_watertight} shows that our NVF outperforms the NDF and GIFS on either the category-agnostic or the category-unseen reconstruction.
For example, we decrease $\sim 10\%$ CD on base classes and $\sim 15\%$ CD on novel classes from NDF.
The qualitative visualization in~\cref{fig:non_watertight} further supports the results, in which NDF and GIFS miss some important details (\eg, seams on benches) and leave a number of holes (\eg, legs of tables, sides of benches), and their surfaces are not as smooth as ours (\eg, the board of watercraft).

\subsection{Cross-domain Cases} 
\label{sec:cross}
Rather than only concentrating on intra-domain performance as most previous reconstruction methods, we also explore the cross-domain ability of our method, in which we
design a cross-domain reconstruction experiment and demonstrate the quantitative results and qualitative visualization respectively in ~\cref{tab:garment} and~\cref{fig:garment}. 
Specifically, we directly deploy our category-agnostic models trained based on ShapeNet base classes to MGN~\cite{mgn} to evaluate its performance. We randomly sample 20 shapes from the 5 categories in MGN and 3,000 points from the raw data as the input.

\cref{tab:garment} indicates that our NVF outperforms other methods by a large margin under all metrics. For example, our method reduces almost $40\%$ and $60\%$ CD respectively from NDF and GIFS. The qualitative visualization in~\cref{fig:garment} also supports this observation, where only our NVF can reconstruct the crinkles of the shirts and pants faithfully.

\begin{table}[t]
\centering
\resizebox{0.48\textwidth}{!}{
    \begin{tabular}{@{}c|c|c|c|c|c|c@{}}
    \hline
     & K & Codebook & CD$\downarrow$ & EMD$\downarrow$ & F1$_{1\times10^{-5}}$ & F1$_{2\times10^{-5}}$ \\
    \hline
     \multirow{3}{*}{Base} & k=8 & \xmark & 0.089 & 1.195 & 74.139 & 89.392	\\
      & k=8 & \cmark & 0.087 & \textbf{1.172} & 75.104 & 90.057  \\
      &  k=16 & \xmark & 0.121 & 1.212 & 73.642 & 88.962 \\
      & k=16 & \cmark & \textbf{0.085} & 1.197 & \textbf{75.372} & \textbf{90.266} \\
     \hline
     \multirow{3}{*}{Novel} & k=8 & \xmark & 0.080 & 1.334 & 78.701 & 90.972 \\
      & k=8 & \cmark& 0.083 & 1.354 & 79.522 & 91.434\\
      & k=16 & \xmark& 0.081 & \textbf{1.329} & 78.800 & 91.084\\
      & k=16 & \cmark & \textbf{0.078} & 1.340 & \textbf{79.723} & \textbf{91.576}\\
     \hline
    \end{tabular}
    }
    \vspace{-3mm}
    \caption{Effect of feature number $K$ and multi-head codebook. The multi-head codebook improves the performance and achieves the best for $K=16$. }
    \label{tab:sampling}
\end{table}

\begin{table}[t]
\centering
\resizebox{0.5\textwidth}{!}{
    \begin{tabular}{@{}c|c|c|c|c|c@{}}
    \hline
     & Backbone & CD$\downarrow$ & EMD$\downarrow$ & F1$_{1\times10^{-5}}$ & F1$_{2\times10^{-5}}$ \\
    \hline
     \multirow{3}{*}{Base} & NDF~\cite{ndf} & 0.099 & 1.372 & 72.425 & 88.754 \\
     \cline{2-6}
     & 3D Conv & 0.092 & \textbf{1.189} & 72.415 & 89.365\\
      & Pointnet++ & 0.088 & 1.198 & 74.156 & 89.418 \\
      & PointTransformer & \textbf{0.085} & 1.197 & \textbf{75.372} & \textbf{90.266} \\
     \hline
     \multirow{3}{*}{Novel} & NDF~\cite{ndf} & 0.093 & 1.532 & 76.162 & 89.977 \\
     \cline{2-6}
     & 3D Conv & 0.090 & 1.342 & 76.155 & 90.462\\
      & Pointnet++ & 0.080 & \textbf{1.339} & 78.444 & 90.850 \\
      & PointTransformer & \textbf{0.078} & 1.340 & \textbf{79.723} & \textbf{91.576}\\
     \hline
    \end{tabular}
    }
    \vspace{-3mm}
    \caption{Comparison of different backbones on non-watertight ShapeNet. The point cloud based backbones outperform the 3D convolution backbone.  PointTransformer performs the best.}
    \label{tab:backbone}
\end{table}

\begin{table}[t]
\centering
\resizebox{0.5\textwidth}{!}{
    \begin{tabular}{@{}c|c|c|c|c@{}}
    \hline
    Methods & Backbone & Codebook & Runtime & Memory \\
    \hline
    NDF~\cite{ndf}& 3D Conv & \xmark & 0.75s & 13.44G \\
    \hline
    \multirow{4}{*}{Ours} & 3D Conv & \xmark & 0.27s & 9.21G \\
     & 3D Conv & \cmark & 0.34s & 9.21G \\
      & PointTransformer & \xmark & 0.29s & 9.28G \\
      & PointTransformer & \cmark & 0.35s & 9.28G \\
     \hline
    \end{tabular}
    }
    \vspace{-3mm}
    \caption{Inference analysis. The runtime and memory are time cost and peak memory during the inference of 200k queries. NVF is more efficient on inference runtime and memory.}
    \label{tab:speed}
    \vspace{-2mm}
\end{table}

\section{Ablation Study}
We conduct comprehensive ablation studies on non-watertight category-agnostic and category-unseen reconstruction (same settings as~\cref{sec:agnostic}.) to evaluate the effectiveness of the proposed modules including multi-head codebook, and the influence of hyper-parameters like the feature number $K$ (i.e., the number of the nearest points in~\cref{sec:FeatureExtraction}) on the results. 
Complexity analysis is also provided to demonstrate the efficiency of our model.

\noindent\textbf{Feature number.} The number of features for query points (determined by the number of nearest points on the point cloud $P$) is an important factor of the model. We report the performance of our network using $k={8,16}$ in~\cref{tab:sampling}, and observe that a larger feature number $k$ yields better results but at the cost of memory. 

\noindent\textbf{Codebook.} We also demonstrate the effectiveness of the introduced codebook mechanism in~\cref{tab:sampling}. The results indicate that using codebook can generally improve the overall performance regarding all metrics. 

\begin{figure}[t]
  \centering
   \includegraphics[width=0.8\linewidth]{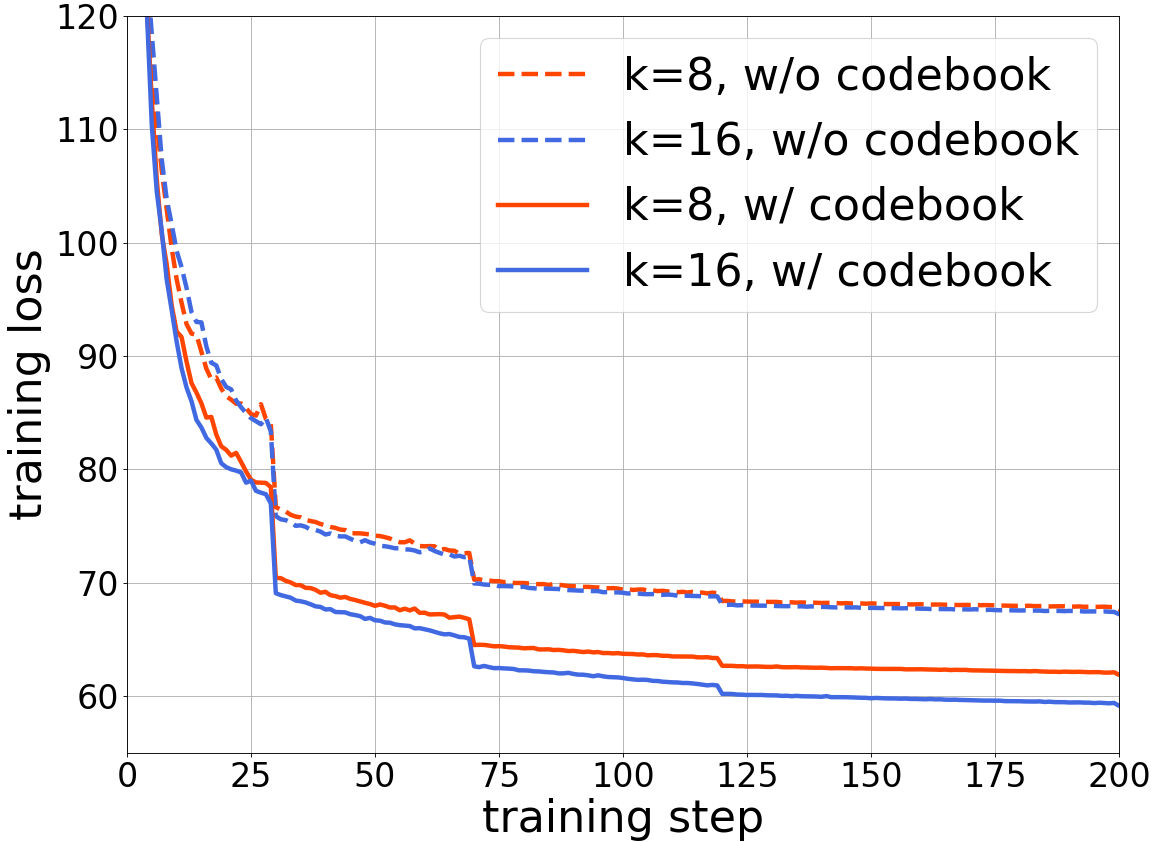}
   \vspace{-2mm}
   \caption{Training curves of models w/ and w/o codebook. The models w/ codebooks converge faster than those w/o  codebooks.}
   \label{fig:loss}
\end{figure}

\noindent\textbf{Training Effectiveness.} We would like to highlight that the introduced multi-head codebook could also work as regularization to reduce the training time. As shown in~\cref{fig:loss}, the loss curves of the models with (w/) codebooks (\ie, solid lines) generally converge faster than their corresponding models without (w/o) codebooks (\ie, dashed lines).

\noindent\textbf{3D Feature Extraction.} We also report our methods with different 3D feature extraction backbones (\ie, 3D convolution~\cite{ifnet,ndf,gifs} , Pointnet++~\cite{pointnet++}, and PointTransformer~\cite{point_transformer}) in ~\cref{tab:backbone}.
The results show that our NVF can already achieve comparable (if not better) results with NDF using the same extraction backbone (\eg, 3D convolution), while ours is more efficient on runtime and memory cost. Using PointTransformer can consistently improve the overall performance. 

\noindent\textbf{Complexity.} We report the inference time and peak GPU memory on the machine with one RTX 3090Ti in~\cref{tab:speed} for obtaining the distance and direction of 200k query points.
The results show that our method only requires $50\%$ inference time and $70\%$ memory compared with NDF when using the same feature extraction backbone. This is due to the cost from the differentiation of distance field in NDF, while our method avoids this issue and introducing multi-head codebook only brings negligible overhead of memory and time as shown in the~\cref{tab:speed}. 

\begin{figure}[t]
  \centering
   \includegraphics[width=0.8\linewidth]{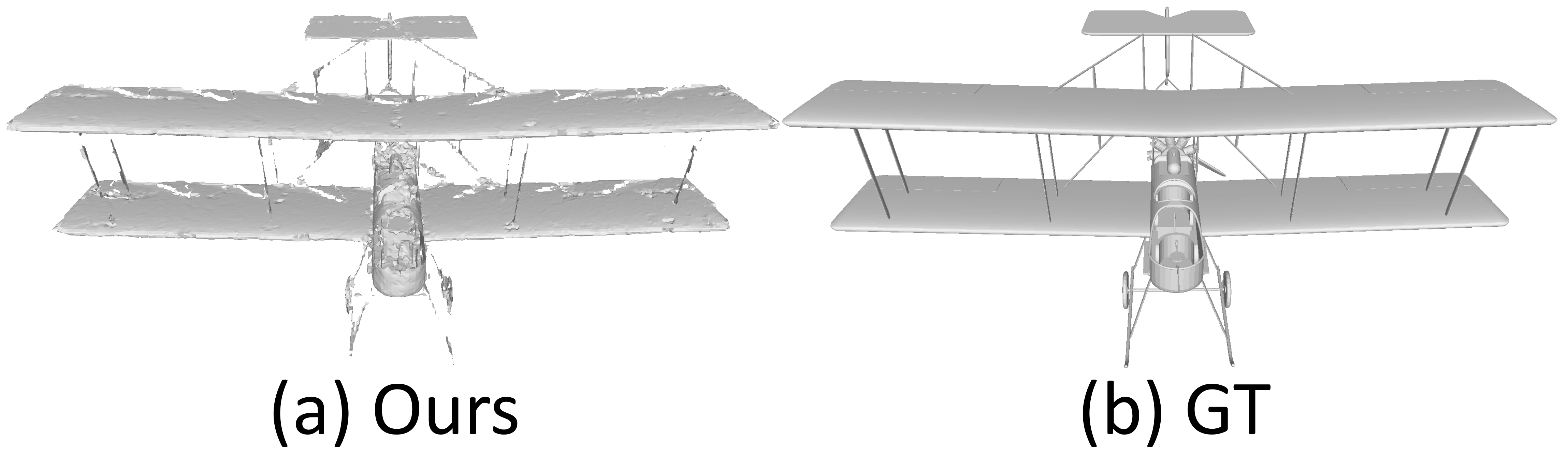}
   \vspace{-2mm}
   \caption{Visualization of one failure case of our model when reconstructing a very thin and complex 3D structure.}
   \label{fig:failure}
    \vspace{-2mm}
\end{figure}

\noindent\textbf{Limitations.} Our method can handle watertight or non-watertight shapes with arbitrary resolution and topology. However, the reconstruction for very thin or complex structures is still far from perfect as visualized in~\cref{fig:failure}.

\section{Conclusion}
In this work, we introduce a novel surface reconstruction representation \textit{NVF}, which leverages the advantages of both the implicit representations and the explicit learning process. It allows the reconstruction of high-quality general object shapes including watertight, non-watertight, and multi-layer shapes. Thanks to our differentiation-free design, we also introduce vector quantization and build a multi-head codebook to improve the generalization of cross-category reconstruction. Experiments demonstrate that our method not only achieves state-of-the-art reconstruction performance for varied topology, but also improves the training and inference efficiency. 

\section{Grant Acknowledgement}
This research is partly supported by Australian Research Council (ARC) DP200103223 and the Ministry of Education, Singapore, under its Academic Research Fund Tier 2 (MOE-T2EP20220-0007) and Tier 1 (RG14/22).

{\small
\bibliographystyle{ieee_fullname}
\bibliography{egbib}
}

\end{document}